\documentclass[a4paper]{article}

\usepackage{ISCSLP2024}
\usepackage{threeparttable}
\usepackage{comment}

\title{An Empirical Study of Retrieval Augmented Generation with Chain-of-Thought}
\name{Yuetong Zhao$^1$, Hongyu Cao$^1$, Xianyu Zhao$^2$, Zhijian Ou$^{1,2}$
\thanks{
Corresponding author: Z. Ou. This work is supported by National Science and Technology Major Project 2023ZD0121401, and partly by Guangxi Science and Technology Project (GUIKEAD23026054).
}
}
\address{
  $^1$Speech Processing and Machine Intelligence (SPMI) Lab, Tsinghua University, China\\
  $^2$TasiTech, China
}
\email{zhaoyuet20@mails.tsinghua.edu.cn, ozj@tsinghua.edu.cn}

\begin{document}

\maketitle

\begin{abstract}
  Since the launch of ChatGPT at the end of 2022, generative dialogue models represented by ChatGPT have quickly become widely used. As user expectations increase, enhancing the capability of generative dialogue models to solve complex problems has become a focal point of current research. This paper delves into the effectiveness of the RAFT (Retrieval Augmented Fine-Tuning) method in improving the performance of Generative dialogue models. RAFT combines chain-of-thought with model supervised fine-tuning (SFT) and retrieval augmented generation (RAG), which significantly enhanced the model's information extraction and logical reasoning abilities. We evaluated the RAFT method across multiple datasets and analysed its performance in various reasoning tasks, including long-form QA and short-form QA tasks, tasks in both Chinese and English, and supportive and comparison reasoning tasks. Notably, it addresses the gaps in previous research regarding long-form QA tasks and Chinese datasets. Moreover, we also evaluate the benefit of the chain-of-thought (CoT) in the RAFT method. This work offers valuable insights for studies focused on enhancing the performance of generative dialogue models.
\end{abstract}

\section{Introduction}

In recent years, with the rapid development of human-computer dialogue, a key technology in this field, generative dialogue models~\cite{zhang2020task,liu2023building},  has shown great potential and wide application prospects. From the early sequence-to-sequence (Seq2Seq)~\cite{Seq2Seq} architecture to recent innovations based on the Transformer~\cite{transformer} model with attention mechanisms, more advanced models are constantly emerging. However, generative dialogue models still confront significant challenges in accuracy, consistency, coherence, security, and resource efficiency. Enhancing their performance is a critical issue that demands attention.

To tackle more complex and diverse NLP tasks, the chain-of-thought (CoT) method~\cite{CoT,BBH,CoT-training,react} has been proposed. Chain-of thought breaks down complex reasoning tasks into multiple intermediate steps that are computed sequentially to obtain the final result. It not only improves the logical consistency of the model’s responses but also enhances user interaction experiences. However, recent studies have shown that chain-of-thought prompting method requires models of \textasciitilde100 billion parameters to fully release their reasoning ability~\cite{CoT}, and thus will have a significant demands on computational resources. 

Retrieval Augmented Generation (RAG)~\cite{RAG} is also a promising method to improve the performance of generative dialogue models~\cite{atlas,RETRO,RAG-luaguage-modeling1,RAG-luaguage-modeling2}. Retrieval augmented generation method enhances the performance and reliability of generative dialogue models by integrating knowledge from external databases. This method not only increases the accuracy and relevance of the generated text but also enables continuous updates of domain-specific knowledge, especially excelling in knowledge-intensive tasks. However, RAG still faces several challenges. Since the performance of retrieval augmented generation depends on the accuracy and efficiency of the retriever, poor-quality or irrelevant retrieval results may negatively impact the generated content. Additionally, how to effectively integrate the retrieved information with the prior knowledge of the model remains a significant challenge.

This paper studies a method that combines the chain-of-thought with retrieval augmented generation for Supervised Fine-Tuning (SFT) small-scale models to optimize their performance in reasoning tasks, which is called RAFT (Retrieval Augmented Fine-Tuning)~\cite{raft}. This method not only avoids the reliance of the chain-of-thought prompting on large-scale models, but also alleviates the hallucination~\cite{hallucination} and maintenance challenges of the knowledge retrieval process in RAG, enhancing the model's ability to extract information and perform logical reasoning. In this work, we provide comprehensive optimization and evaluation of RAFT method across different types of reasoning tasks, including short-form QA and long-form QA, English tasks and Chinese tasks, bridge type and comparison tasks, particularly focusing on long-form QA and Chinese datasets. In addition, we evaluated the benefits of the chain-of-thought in the RAFT method and conducted a detailed analysis of the performance across various type of tasks above.

\begin{figure*}
  \centering
  \includegraphics[width=1\linewidth]{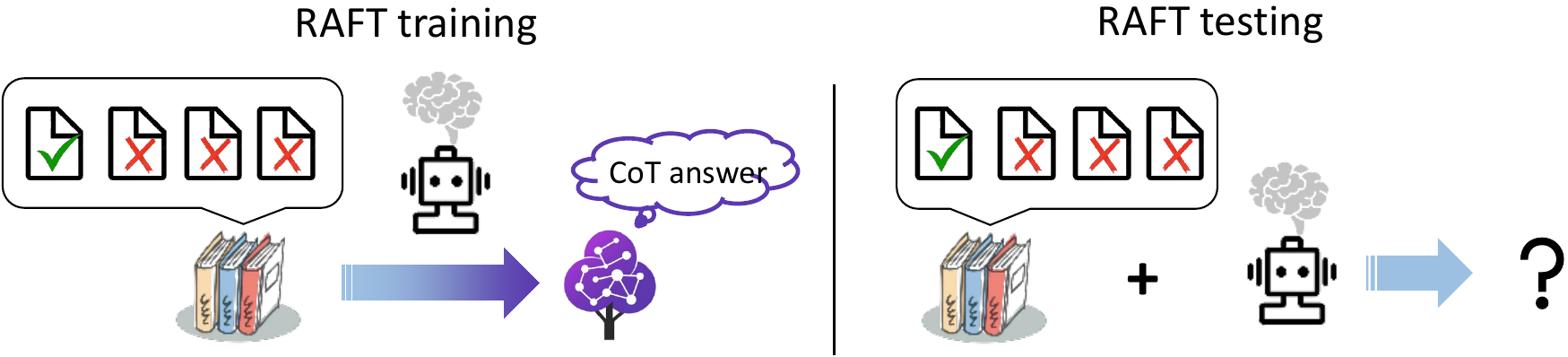}
  \caption{Overview of RAFT}
  \label{fig:3}
\end{figure*}

\begin{figure*}
  \centering
  \includegraphics[width=1\linewidth]{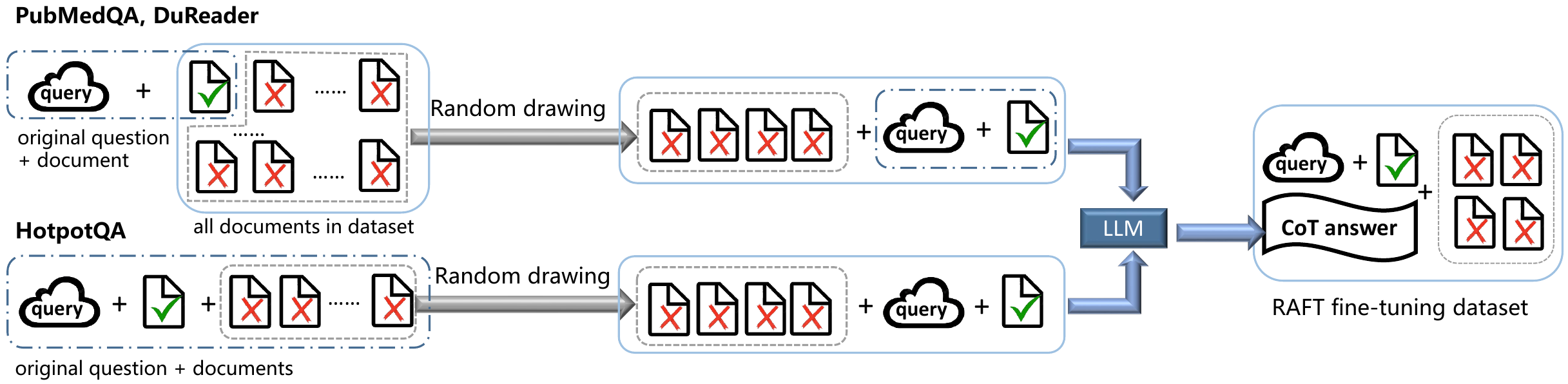}
  \caption{Construction of RAFT fine-tuning dataset}
  \label{fig:4}
\end{figure*}

\begin{figure}
  \centering
  \includegraphics[width=1\linewidth]{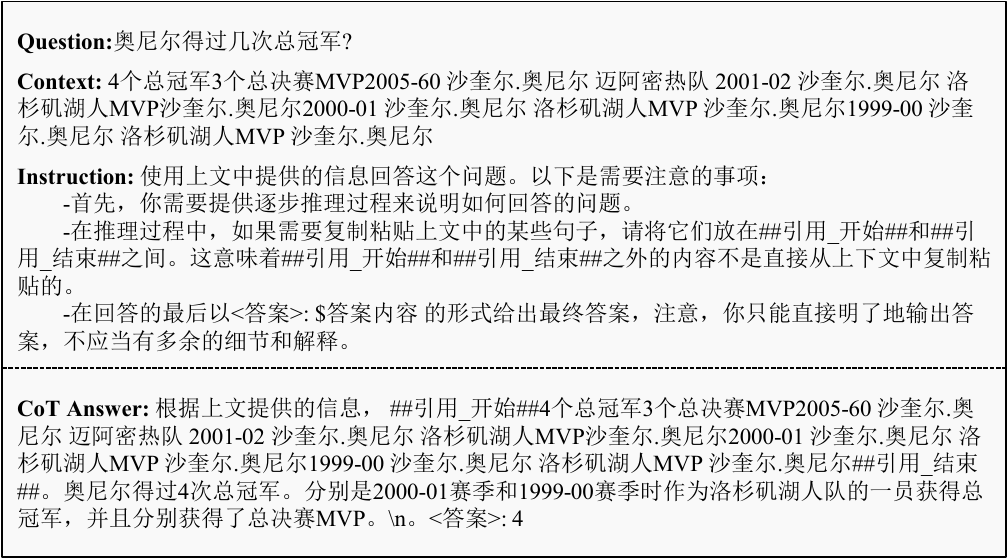}
  \caption{Examples of Chinese chain-of-thought style response generation process via GPT-3.5 in RAFT.}
  \label{fig:10}
\end{figure}

\section{Method}
\subsection{RAFT Finetuning}
RAFT~\cite{raft} is derived from RAG+SFT, which combines retrieval augmented generation (RAG) and Supervised Fine-Tuning (SFT). For better understanding, let us make an analogy between these modeling techniques in generative dialogue models and various kinds of examinations faced by human.

For supervised fine-tuning, the pre-trained language model is fine-tuned for a specific task by introducing a labeled dataset tailored for the task. Supervised fine-tuning is similar to a closed-book exam taken after class, where students answer questions using only the problem-solving methods learned in class without any reference materials. 

For the retrieval augmented generation method, the RAG model uses input prompts as query keywords to retrieve relevant documents. These retrieved contents are added to the model’s input, and the model generates responses based on the augmented input. In the examination analogy, this method can be regarded as finding relevant passages from the open-book knowledge according to the question and reasoning the answer.

The RAFT method combines retrieval augmented generation and supervised fine-tuning, as well as incorporating the idea of chain-of-thought. This is akin to training the model to compute results from relevant information before taking an exam. Consequently, during an open-book exam, the model can deduce correct answers more quickly and accurately using the reference materials. \emph{In summary, the RAFT method has two key features.} First, in addition to the oracle documents, irrelevant distractor documents are also included in the reference documents to improve model's robustness against irrelevant information retrieved during the retrieval process. Second, chain-of-thought style responses are used as the target text in the fine-tuning dataset rather than plain short answers to improve model's reasoning capability. To be specific, each data in RAFT dataset contains a question ($Q$), several distractor documents ($D_k$), a oracle document containing the effective information to answer the question ($D^*$), and a chain-of-thought style response ($A^*$) generated from the oracle document ($D^*$). Figure~\ref{fig:3} shows overview of RAFT.

\begin{equation}
\setlength{\abovedisplayskip}{2pt plus 1pt minus 1pt}
\setlength{\belowdisplayskip}{2pt plus 1pt minus 1pt}
      RAFT\ training: Q+D^*+D_1+D_2+\dots+D_k\rightarrow A^*
\end{equation}

\begin{figure}
  \centering
  \includegraphics[width=1\linewidth]{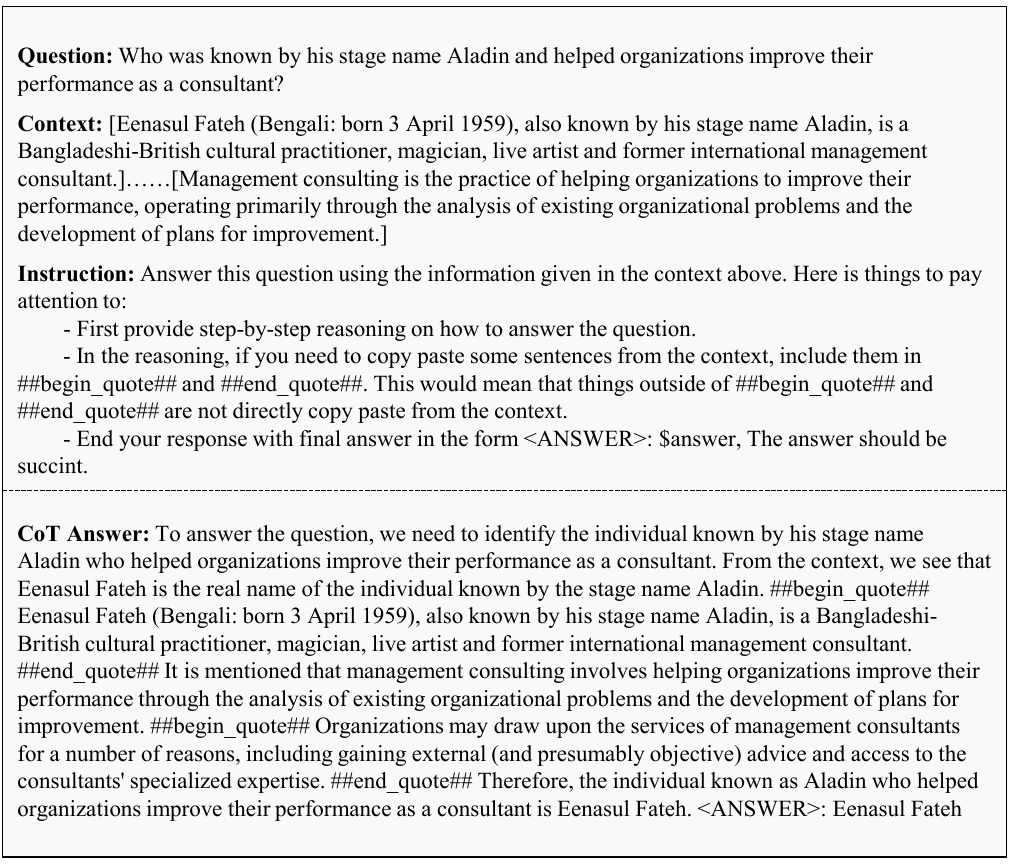}
  \caption{Examples of English chain-of-thought style response generation process via GPT-3.5 in RAFT.}
  \label{fig:5}
\end{figure}

\subsection{Dataset Construction}
Figure~\ref{fig:4} shows our RAFT fine-tuning dataset construction process. 
 In order to make the datasets tailored to RAFT fine-tuning, we use two methods to process open-source datasets. When dealing with a dataset where a question corresponds to several reference documents (include oracle documents and distractor documents), we use the first method: For each question, we extract all the oracle documents from the question's corresponding documents, then randomly select a specified number of documents from the remaining corresponding documents as the distractor documents. When dealing with a dataset where a question corresponds to only one oracle document, we use the second method: For each question, we take its corresponding document as oracle document and randomly select a specified number of documents from \textit{other} questions' reference documents as the question's distractor documents. In this study, the dataset HotpotQA \cite{hotpotqa} was processed using the first method, while the datasets like PubMedQA \cite{pubmedqa} and DuReader\_robust \cite{dureader} were processed using the second method. In our RAFT experiments, we use four distractor documents for each question.

After selecting the oracle and distractor documents, we used GPT-3.5 with to generate chain-of-thought style response. We require the model to generate a chain-of-thought reasoning process based on the input question as well as its corresponding oracle documents. During this reasoning process, the model are prompted to cite the referenced content from the oracle documents and provide a final answer separately at the end. Figure~\ref{fig:10} and figure~\ref{fig:5} show our CoT answer generation process via GPT-3.5 in Chinese and English respectively.

\section{Experiment Setup}
\subsection{Datasets}
\begin{itemize}
\item HotpotQA \cite{hotpotqa}: HotpotQA dataset contains 113,000 multi-hop reasoning question-answer pairs from Wikipedia. It includes two types of QA tasks: bridge and comparison. Bridge QA tasks require the model to find relevant information from multiple reference documents to provide an answer, while comparison QA tasks require the model to compare multiple entities or events. Each data item includes a question, several reference documents, and a short answer.

\item PubMedQA \cite{pubmedqa}: A biomedical question-answering dataset. It extracts data from PubMed abstracts and answers research questions based on these abstracts. Answers are presented in the form of "yes/no/maybe." Each data item consists of a question, a reference document, a long answer, and a short answer.

\item DuReader\_robust \cite{dureader}: DuReader\_robust is a Chinese dataset used to evaluate the robustness and generalization ability of model's reading comprehension function. Each data item includes a question, a reference document, and a short answer. All datas items are sourced from Baidu users' search queries and responses.
\end{itemize}

\subsection{Baselines}
In this study, we evaluated the Chinese dataset DuReader\_robust using Qwen-1.5-7B-chat \cite{qwen} and the English datasets, HotpotQA and PubMedQA, using LLaMA2-7B-chat \cite{llama,llama2}.

\begin{itemize}
\item LLaMA2-7B-chat / Qwen-1.5-7B-chat + zero-shot prompting: Provide the model with clear instructions and the question it needs to answer, without providing any external reference documents, and require the model to generate an answer.

\item LLaMA2-7B-chat / Qwen-1.5-7B-chat + RAG: Provide the model with instructions and the question, supplemented with external reference documents, and require the model to derive an answer using the content in these reference documents.

\item DSF (Domain Specific Finetuning) + zero-shot prompting: For each dataset standard supervised fine-tuning is performed without reference documents, using the question as the input text and the answer as the target text for fine-tuning. The fine-tuned model is then given the question and instructions and required to respond without referencing external documents.

\item DSF + RAG: Standard supervised fine-tuning is performed without reference documents for each dataset, but during testing, the fine-tuned model is supplemented with reference documents for the question. The model is required to derive the answer using external knowledge.
\end{itemize}

\subsection{Evaluation Method}
In our experiments, we primarily use F1 score and EM score (Exact Match) to evaluate the performance of the models. We standardize the answers by normalizing the answer text through several steps, including converting all text to lowercase, removing punctuation, removing articles (a, an, the), and standardizing spaces, which ensure that the answers are more uniform in format \cite{DuReader-code,HotpotQA-code}. Subsequently, the standardized answers are used to calculate their EM scores and F1 scores.
\section{Experiment Result}

\begin{table}[]
\setlength{\tabcolsep}{3pt}
\centering
\begin{threeparttable}
\caption{Evaluation in EM score}
\begin{tabular}{cccc}
\toprule
                         & PubMedQA & HotpotQA$_{[Oracle]}$\ & HotpotQA \\ \midrule
zero-shot       & 50.50    & 15.06    & 15.06   \\
RAG                       & 56.42    & 12.07    & 8.72    \\
DSF + zero-shot & 53.91    & 20.04    & 20.04   \\
DSF + RAG                 & 71.71    & 45.26    & 27.40   \\ \midrule
RAFT w.o. CoT    & 54.80    & 52.38    & 28.74   \\ \midrule
RAFT                      & 74.36    & 54.20    & 39.48   \\ \bottomrule
\end{tabular}
\label{table1}
\end{threeparttable}
\end{table}

\begin{table*}[]
\centering
\begin{threeparttable}
\caption{Evaluation in F1 score}
\begin{tabular}{ccccc}
\toprule
                         & PubMedQA$_{[long]}$ & HotpotQA$_{[Oracle]}$ & HotpotQA & DuReader \\ \midrule
zero-shot       & 1.09  & 22.63    & 22.63   & 13.47    \\
RAG                       & 3.05  & 25.05    & 18.39   & 26.06    \\
DSF + zero-shot & 7.95  & 27.63    & 27.63   & 20.90    \\
DSF + RAG                 & 10.68 & 58.67    & 34.52   & 39.91    \\ \midrule
RAFT w.o. CoT    & ——    & 64.47    & 37.48   & 42.25    \\ \midrule
RAFT                      & 14.09 & 67.83    & 51.33   & 57.81    \\ \bottomrule
\end{tabular}
\label{table2}
\end{threeparttable}
\end{table*}

We compared the performance of the models using the RAFT method and the baselines. Table~\ref{table1} and Table~\ref{table2} show the results for the EM score and F1 score respectively. In the HotpotQA$_{[Oracle]}$ experiment group, only oracle documents were provided as references for the model in the RAG experiments. For all other groups, distractor documents were included alongside the reference documents in the RAG experiments.

From the experimental results, we can see that the RAFT method consistently outperforms four baseline methods across all datasets, demonstrating superior information extraction and complex problem reasoning capabilities in the models fine-tuned with RAFT method. On the HotpotQA dataset, the RAFT method (with CoT) achieved a performance gain of 42.13\% in EM score and 42.78\% in F1 score over the plain RAG baseline (without using DSF model) experiments. Even with the inclusion of distractor documents, it still achieved gains of 30.76\% in EM score and 32.94\% in F1 score. Furthermore, we observed that although the scores for RAFT degrade with the addition of distractor documents in the experiments (comparing the table columns corresponding to HotpotQA$_{[Oracle]}$ and HotpotQA), it achieved a higher performance gain over the DSF+RAG baseline. This indicates that the RAFT method can significantly enhance the model's robustness in the retrieval process in RAG.

Before fine-tuning, the model's performance was poor, regardless of whether RAG was included or not. Fine-tuning the model for specific domains, i.e., DSF, can significantly improve its performance by aligning model outputs with the answering patterns of those domains. Through the RAFT method (with CoT), the model not only learned specific domain answering patterns but also significantly improved its ability to extract effective information from complex data.

\subsection{Long-form QA Evaluation}
Since the "yes/no" QA of PubMedQA and QA of HotpotQA are both short-form, we also assessed the long-form QA in dataset PubMedQA. The experiment results are shown in Table~\ref{table2} under the PubMedQA$_{[long]}$ group. The results in F1 score of long-form QA indicate that RAFT method brought about a 13\% performance improvement for long-answer questions over zero-shot prompting baseline. However, compared to the DSF+RAG baseline, the performance gain was less prominent than for the short-form QA. This is because the content of long answers is more focused on induction and summarization, rather than definitive results derived from reasoning, as is common with short answers. The study of long-form QA with chain-of-thought needs further exploration. 

\subsection{Chinese Dataset Evaluation}
We also conducted evaluation on DuReader\_robust to assess the effectiveness of the RAFT method on the Chinese datasets. Since the questions in this dataset heavily rely on information from reference documents, the gain brought by the use of DSF is only 7.43\% over the zero-shot prompting baseline (in Table~\ref{table2} comparing the 'zero-shot' and 'DSF+zero-shot' rows in the DuReader group). In this case, the use of RAG to supplement reference documents with the question is more effective, which obtains a 12.59\% performance gain over the zero-shot baseline. After RAFT fine-tuning, the model's ability to extract and process information, as well as its reasoning capability can be significantly improved. It achieves 44.34\% and 19.9\% performance gain in F1 score over zero-shot prompting baseline and DSF+RAG baseline respectively. These results demonstrate that the RAFT method performs exceptionally well on both English and Chinese datasets.

\subsection{Performance Across Different Types of Reasoning Tasks by RAFT}

We evaluated the RAFT method separately on bridge-type QA and comparison-type QA in HotpotQA dataset, as shown in Table~\ref{table3}. The results indicate that RAFT performs better on comparison-type questions. This is likely because comparison-type questions typically involve comparing features between two or more entities, which can rely on direct information retrieval and simple comparison operations. In contrast, bridge-type questions often require the model to extract relevant information from multiple documents, involving longer reasoning chains and multiple intermediate steps so it demands a higher level of understanding and reasoning ability from the model.

\begin{table}[]
\centering
\caption{Performance gains across different types of reasoning tasks by RAFT}
\begin{tabular}{ccc}
\toprule
     & bridge & comparison \\ \midrule
RAFT-EM score & 36.25  & 50.72      \\
RAFT-F1 score & 48.80  & 60.11      \\ \bottomrule
\end{tabular}
\label{table3}
\end{table}

\subsection{Effect of CoT}
To evaluate the benefit of the chain-of-thought (CoT) in the RAFT method, we conducted an ablation experiment (RAFT w.o. CoT). In this experiment, we removed the chain-of-thought style response from the RAFT training dataset and only included the final answer for each question as target text in the fine-tuning process. Comparing the HotpotQA dataset tested with only oracle documents for RAG and the one tested with distractor documents, the CoT method achieved more significant performance gains in the latter setting. This demonstrates that CoT can obtain more considerable benefit in the face of more complex knowledge and more serious information noise. Moreover, the performance of RAFT was consistently superior to the performance of RAFT without CoT across various datasets. Therefore, adding CoT effectively guides the model the correct information from complex input and enhances the model's logical rigor and accuracy.

\section{Conclusion}
In this study, we evaluated the RAFT method across multiple datasets, addressing the gaps in previous research regarding long-form QA and Chinese datasets. The results indicate that the RAFT method combined with CoT not only improves the models' ability to robustly extract and process information in the face of noise, but also enhances their logical reasoning ability in reasoning tasks. Significant performance gains were observed in evaluations on both English and Chinese datasets, as well as on long-form QA and short-form QA. Additionally, we conducted an ablation experiment where we removed the chain-of-thought style response from the RAFT training dataset to fine-tune the model. This experiment verifies the critical role of the chain-of-thought in enhancing the performance of generative dialogue models.

\begin{comment}
\begin{table*}[]
\begin{tabular}{@{}cccccc@{}}
\toprule
                         & PubMedQA$_{[long]}$ & PubMedQA     & HotPotQA$_{[Oracle]}$     & HotPotQA     & Dureader \\ \midrule
Few-shot prompting       & 1.09     & 50.50 (50.50) & 22.63 (15.06) & 22.63 (15.06) & 13.47    \\
RAG                      & 3.05     & 56.42 (56.42) & 25.05 (12.07) & 18.39 (8.72)  & 26.06    \\
DSF + Few-shot prompting & 7.95     & 53.91 (53.91) & 27.63 (20.04) & 27.63 (20.04) & 20.90    \\
DSF + RAG                & 10.68    & 71.71 (71.71) & 58.67 (45.26) & 34.52 (27.40) & 39.91    \\ \midrule
RAFT w.o CoT             & ——       & 54.80 (54.80) & 64.47 (52.38) & 37.48 (28.74) & 42.25    \\ \midrule
RAFT                     & \underline{14.09}    & \underline{74.36 (74.36)} & \underline{67.83 (54.20)} & \underline{51.33 (49.48)} & \underline{57.81}    \\ \bottomrule
\end{tabular}
\end{table*}
\end{comment}
\clearpage

\bibliographystyle{IEEEtran}

\bibliography{mybib}

% \begin{thebibliography}{9}
% \bibitem[1]{Davis80-COP}
%   S.\ B.\ Davis and P.\ Mermelstein,
%   ``Comparison of parametric representation for monosyllabic word recognition in continuously spoken sentences,''
%   \textit{IEEE Transactions on Acoustics, Speech and Signal Processing}, vol.~28, no.~4, pp.~357--366, 1980.
% \bibitem[2]{Rabiner89-ATO}
%   L.\ R.\ Rabiner,
%   ``A tutorial on hidden Markov models and selected applications in speech recognition,''
%   \textit{Proceedings of the IEEE}, vol.~77, no.~2, pp.~257-286, 1989.
% \bibitem[3]{Hastie09-TEO}
%   T.\ Hastie, R.\ Tibshirani, and J.\ Friedman,
%   \textit{The Elements of Statistical Learning -- Data Mining, Inference, and Prediction}.
%   New York: Springer, 2009.
% \bibitem[4]{YourName17-XXX}
%   F.\ Lastname1, F.\ Lastname2, and F.\ Lastname3,
%   ``Title of your ISCSLP 2024 publication,''
%   in \textit{ISCSLP 2024 -- 23\textsuperscript{rd} Annual Conference of the International Speech Communication Association, September 18-22, Incheon, Korea, Proceedings, Proceedings}, 2024, pp.~100--104.
% \end{thebibliography}

\end{document}